\documentclass[aps,preprint,superscriptaddress]{revtex4}

\usepackage{hyperref}
\hypersetup{hidelinks}

\usepackage{graphicx}
\usepackage{dcolumn}
\usepackage{bm}  
\usepackage{amssymb}
\usepackage{amsmath}
\usepackage[dvipsnames]{xcolor}
\usepackage{ulem}
\begin{document}
\title{Entropy Regularized Reinforcement Learning\\ Using Large Deviation Theory}
\author{Argenis Arriojas}
\email{arriojasmaldonado001@umb.edu}
\affiliation{Department of Physics, University of Massachusetts Boston, Boston MA 02125, USA}
\author{Jacob Adamczyk}
\affiliation{Department of Physics, University of Massachusetts Boston, Boston MA 02125, USA}
\author{Stas Tiomkin}
\affiliation{Department of Computer Engineering, San Jose State University, San Jose CA 95192, USA}
\author{Rahul V. Kulkarni}
\email{rahul.kulkarni@umb.edu}
\affiliation{Department of Physics, University of Massachusetts Boston, Boston MA 02125, USA}
\date{\today}

\begin{abstract}
Reinforcement learning (RL) is an important field of research in machine learning that is increasingly being applied to complex optimization problems in physics. In parallel, concepts from physics have contributed to important advances in RL with developments such as entropy-regularized RL. While these developments have led to advances in both fields, obtaining analytical solutions for optimization in entropy-regularized RL is currently an open problem. In this paper, we establish a mapping between entropy-regularized RL and research in non-equilibrium statistical mechanics focusing on Markovian processes conditioned on rare events. In the long-time limit, we apply approaches from large deviation theory to derive exact analytical results for the optimal policy and optimal dynamics in Markov Decision Process (MDP) models of reinforcement learning. The results obtained lead to a novel analytical and computational framework for entropy-regularized RL which is validated by simulations. The mapping established in this work connects current research in reinforcement learning and non-equilibrium statistical mechanics, thereby opening new avenues for the application of analytical and computational approaches from one field to cutting-edge problems in the other.

\end{abstract}

\maketitle

The combination of machine learning approaches with concepts and tools from physics has given rise to significant developments in current research ~\cite{sarma2019machine}. Concepts derived from statistical mechanics have led to important applications in machine learning~\cite{mehta2019high} and recent work has further highlighted the importance of building bridges between the two disciplines~\cite{mehta2014exact,lin2017does,bahri2020statistical}. On the flip side, machine learning approaches such as reinforcement learning (RL) are increasingly being used to address complex optimization problems in diverse fields of physics, ranging from quantum computing and quantum control to adaptive optics~\cite{ostaszewski2021reinforcement,barr2020quantum,Bolens2021Sep,Zhang2020Oct,Borah2021Nov,Nousiainen2021May}.
While RL approaches are now being widely applied in physics research, there has been less emphasis on using insights and approaches from  physics to address open problems in RL.
The development of such approaches can lead to important discoveries in RL research as well as provide avenues for the development of novel RL algorithms to solve a diverse range of problems in physics~\cite{barr2020quantum,rose2021reinforcement}.

While the connections of machine learning to equilibrium statistical mechanics are well established~\cite{mehta2019high}, the interface with non-equilibrium statistical mechanics (NESM) is less explored. Recent work has addressed this gap by developing machine learning approaches with applications to NESM. For example, graph neural network models for estimation of the scaled cumulant generating function for observables in dynamical systems have been developed~\cite{Yan2022PhysicsInformed} and evolutionary RL approaches have been used to calculate the likelihood of dynamical large deviations~\cite{Whitelam2020Evolutionary}. While RL approaches are thus starting to be applied to study systems of interest in NESM, it is also of interest to explore if insights from NESM can be used to obtain new insights into RL.
An example of the latter case arises when considering RL problems that involve optimization over system trajectories with entropy-based regularization~\cite{Levine2018May,Kappen2012May}. This framework, termed maximum entropy RL, or more generally entropy-regularized RL, allows the optimal control problem in RL to be recast as a problem in Bayesian inference. This `control-as-inference' approach involves the introduction of optimality variables such that the {\em posterior} trajectory distribution, conditioned on optimality, provides the solution to the optimal control problem~\cite{todorov2008general,rawlik2012stochastic,Kappen2012May,Levine2018May}. While this framework has led to several advances, there are open questions relating to the derivation of analytical results that characterize the optimal dynamics.

Recent research in NESM using large deviation theory has developed a framework for analyzing Markovian processes conditioned on rare events~\cite{garrahan2009first,jack2010large,Chetrite2013Sep,chetrite2015nonequilibrium,chetrite2015variational}.
In this framework, a generalization of the Doob $h$-transform~\cite{Miller1961,Simon2009,chetrite2015nonequilibrium} is used to determine the {\em driven} process: a conditioning-free Markovian process that has the same statistics as the original Markovian process conditioned on a rare event. 
Similar derivations of the driven or controlled processes have been obtained in previous work using a maximum entropy approach for characterizing non-equilibrium steady states~\cite{Evans2004,Evans2005,Simha2008,Jack2016}.
The connection of this framework to RL can be seen by noting that the goal in entropy-regularized RL is to derive the posterior trajectory distribution conditioned on optimality and, in the long-time limit, optimality of the trajectory is a {\em rare event} for the original dynamics. This commonality of conditioning on rare events suggests that approaches and results from NESM can be used to characterize the optimal control policy for entropy-regularized RL problems. Indeed, recent work has explored connections between entropy-regularized RL and rare trajectory sampling and applied it to a range of problems in physics~\cite{rose2021reinforcement,das2021reinforcement}, however an explicit characterization of the optimal controlled processes for general entropy-regularized RL problems has not been derived to date.

In this paper, we develop a mapping between MDP-based entropy-regularized RL and Markovian processes conditioned on rare events in the long-time limit. Using approaches from large deviation theory, we derive exact analytical expressions characterizing trajectory distributions conditioned on optimality. Interestingly, our derivation of these results show how the generalized Doob $h$-transform arises naturally from Bayesian inference applied to trajectory distributions. The results obtained lead to analytical expressions for the optimal policy and optimal dynamics in entropy-regularized RL which are validated using simulations. The connections established in this work also lead to a novel approach for model-free RL and provide new avenues for research focusing on the intersection of RL and physics. 
Specifically, the mapping developed in this work connects RL-based optimization to the estimation of dynamical free energy in NESM~\cite{garrahan2009first}, thus paving the way for the use of approaches such as deep RL to estimate dynamical free energies in nonequilibrium physics.

\section{Markov Decision Process Framework}
In the following, we provide an overview of the standard Markov Decision Process (MDP) framework for reinforcement learning. To introduce the formalism, we focus on the finite horizon, undiscounted case with horizon $N$~\cite{Levine2018May}. 
Consider a Markov chain with states represented by tuples $(s,a)$, where $s$ is an agent's current state and $a$ is an action taken while in state $s$. The probability that the agent transitions to state $s'$ after taking action $a$ is denoted by $p(s'|s,a)$. The choice of action $a$ given the agent's current state $s$ is drawn from a policy $\pi(a|s)$ and the corresponding reward collected by the agent is given by the reward function $r(s,a)$. 

With the above representation, we can now define probability distributions over trajectories $\tau:= \{(s_1, a_1),\dots,(s_{N}, a_{N})\}$ that are generated by the policy $\pi(a|s)$ and transition probabilities $p(s'|s,a)$.
Let $p(s_1)$, $\pi(a|s)$ and $p(s'|s,a)$ denote prior distributions for the initial state, policy, and transition dynamics respectively. The corresponding probability distribution for \emph{uncontrolled} trajectories is given by
\begin{equation}
    p(\tau) = p(s_1)\prod_{t=1}^{N} p(s_{t+1}|s_t,a_t)\pi(a_t|s_t).\label{eq:prior_prob_trajectory}
\end{equation}
The prior distribution for the transition dynamics  corresponds to the system's uncontrolled transition dynamics. In the special case of MaxEnt RL, the prior policy is chosen as the uninformative prior, i.e., a uniform distribution over actions.

We now consider the probability distribution for \emph{controlled} trajectories that is generated by a specific policy $\pi_c(a|s)$ and transition dynamics $p_c(s'|s,a)$ that may, in general, be different from the uncontrolled prior distributions. The probability distribution for controlled trajectories is given by
\begin{equation}
    p_c(\tau) = p_c(s_1)\prod_{t=1}^{N} p_c(s_{t+1}|s_t,a_t)\pi_c(a_t|s_t).\label{eq:prob_trajectory}
\end{equation}

The objective in standard RL is to find the policy $\pi^*(a|s)$ that maximizes the total expected reward. Let $R_\tau=\sum_{t=1}^{N} r(s_t,a_t)$ denote the total reward accumulated over a trajectory $\tau$. Correspondingly, the optimal policy $\pi^*(a|s)$ is given by
\begin{equation}
    \pi^*(a|s) = \arg\max_{\pi_c} \mathbb{E}_{p_c(\tau)} \left[ R_\tau\right].
\end{equation}
In entropy-regularized RL, the goal is to determine the decomposition (Eq.~\ref{eq:prob_trajectory}) for the optimally controlled trajectory distribution $p_c(\tau)$ that maximizes the objective function
\begin{equation}
\mathbb{E}_{p_c(\tau)} [R_\tau] - \frac{1}{\beta} \mathcal{H}(p_c(\tau)||p(\tau)),\label{eq:objective_function}
\end{equation}
where $\beta$ is a regularization parameter corresponding to the inverse temperature. We can see that, in entropy-regularized RL, the standard RL objective function is
augmented to include a regularization term $- \frac{1}{\beta} \mathcal{H}(p_c(\tau)||p(\tau))$. This term corresponds to the relative entropy between the controlled trajectory distribution $p_c(\tau)$ and the prior trajectory distribution $p(\tau)$, and is given by the Kullback-Leibler divergence
\begin{equation*}
    \mathcal{H}(p_c(\tau)||p(\tau)) = \sum_\tau p_c(\tau) \log \frac{p_c(\tau)}{p(\tau)}.
\end{equation*}
This regularization process naturally yields stochastic optimal policies, a desirable feature providing robustness to changes in the problem's dynamics. The role of the $\beta$ parameter is then to regulate the trade-off between obtaining a single ``greedy'' optimal solution, instead of a collection of solutions that offer possibly lower returns in exchange for their robustness.

The preceding generalization of standard RL allows to recast the optimal control problem as an inference problem~\cite{Levine2018May}. This control-as-inference approach involves the introduction of optimality variables $\mathcal{O}_t$ defined such that
\begin{equation}
\label{eq:optimality_var_single_step}
    p(\mathcal{O}_t = 1 | s_t, a_t) = \exp(\beta r(s_t,a_t)),
\end{equation}
The binary random variable $\mathcal{O}_t$ represents the probability that the trajectory is optimal at time step $t$. The purpose of this definition is that the {\em posterior} trajectory distribution, obtained by conditioning on $\mathcal{O}_t =1$ for all $t$, exactly corresponds to the trajectory distribution generated by optimal control. The optimal control problem in entropy-regularized RL thus becomes equivalent to a problem in Bayesian inference. 

Let $\mathcal{O}_{1:N}$ define the event for which all steps in a trajectory $\tau$ are optimal, i.e.
$\mathcal{O}_{1:N} \doteq\bigcap_{i=1}^N \left( \mathcal{O}_i = 1 \right)$. To make connections to the ``statistical mechanics of trajectories'' formalism in NESM~\cite{garrahan2009first}, let us denote by $E_\tau = -R_\tau$ the accumulated \textit{energetic cost} for a trajectory $\tau$. From Bayes' theorem, it follows that the posterior probability distribution for trajectories, conditioned on $\mathcal{O}_{1:N}$,  is given by
\begin{equation}
    p(\tau|\mathcal{O}_{1:N}) 
     = \frac{{p(\tau)} e^{-\beta E_\tau}}{\sum_\tau {p(\tau)} e^{-\beta E_\tau}}.
    \label{eq:bayes_rhs}
\end{equation}

From the inference perspective, the central problem in entropy-regularized RL is now to determine the posterior distributions for the policy, dynamics and initial state, {\em conditioned on optimality}. As noted, these posterior distributions correspond to the solution of the optimal control problem in entropy-regularized RL.

In many practical RL problems, control of system dynamics and initial state distributions is unfeasible. In these cases, the posterior dynamics and initial state distributions must be constrained to exactly match the prior dynamics and initial state distributions and the optimization is carried out by varying the policy alone. We will refer to this approach as the {\em constrained} optimization approach to entropy-regularized RL.
In the constrained optimization problem, the agent only has control over the policy.
The optimal trajectory distribution for the constrained problem can therefore be decomposed as
\begin{align}
    p(\tau|\mathcal{O}_{1:N}) 
    &= p(s_1) \prod_{t=1}^{N} p(s_{t+1}|s_t,a_t)  \pi(a_{t}|s_{t},\mathcal{O}_{1:N}).
    \label{eq:bayes_lhs_constrained}
\end{align}

The preceding (constrained) problem formulation is to be contrasted with the unconstrained optimization problem, where the agent also has control over the transition dynamics and initial state distributions. In this case, the optimal trajectory distribution can be decomposed as \cite{Levine2018May}
\begin{align}
    p(\tau|\mathcal{O}_{1:N}) 
    &= p(s_1|\mathcal{O}_{1:N}) \prod_{t=1}^{N} p(s_{t+1}|s_t,a_t,\mathcal{O}_{1:N}) \nonumber\\
    &\hspace{8em} \times  \pi(a_{t}|s_{t},\mathcal{O}_{1:N}).
    \label{eq:bayes_lhs}
\end{align}
In the remainder of the paper, unless otherwise stated, we will focus on the solution of the unconstrained optimization problem in entropy-regularized RL, where the transition dynamics and the initial state distribution are optimized along with the policy. We note that the framework developed in this work also leads to the solution of the constrained optimization problem, which will be shown elsewhere.

\section{Solution using large deviation theory}
We now proceed to provide an analytical solution to the central problem of entropy-regularized RL in the long-time limit.
Without loss of generality \cite{Levine2018May}, we consider reward functions such that the maximum reward is set to zero and we have $r(s,a)~\leq~0$ for all$~s,a$. In this case, Eq.~\ref{eq:optimality_var_single_step} indicates that in the long-time limit, optimality of the entire trajectory is a rare event and the problem of determining the posterior policy and dynamics corresponds to conditioning on such a rare event. 
Research in NESM~\cite{jack2010large,chetrite2015nonequilibrium} has developed a framework for characterizing Markovian processes conditioned on rare events. In the following, we show how this framework leads to analytical expressions for quantities of interest in entropy-regularized RL. We note that the core of the derivation runs parallel to previous results deriving the Doob $h$-transform in discrete-time Markov chains ~\cite{rogers2000diffusions,levin2017markov,meleard2012quasi,van2013quasi}. In the following, our focus is on applying this framework to obtain new results for entropy-regularized RL.  

Let $z=(s,a)$, $z'=(s',a')$ denote two consecutive state-action tuples. We can combine the system dynamics $p(s'|s,a)$ with the fixed prior policy $\pi(a'|s')$ to compose the corresponding transition matrix for the discrete time Markov chain
\begin{equation}
    P_{ji}= p(z'=j|z=i) = p(s'|s,a)\pi(a'|s').\label{eq:transition_matrix}
\end{equation}

Based on the connection to large deviation theory~\cite{touchette2011basic}, let us define the \textit{tilted transition matrix}
\begin{equation}
    \widetilde{P}_{ji}=P_{ji}e^{\beta r_i},\label{eq:tilted_transition_matrix}
\end{equation}
where $r_i=r(z=i)=r(s,a)$ denotes the reward associated to the tuple $(s,a)$. Note that the tilted matrix is not a stochastic matrix and thus it cannot be interpreted as a transition matrix for a Markov chain that conserves probability. To address this issue, we introduce an additional absorbing state for the agent such that the extended transition matrix $\overline{P}$ (as defined below) {\em is} a stochastic matrix:
\begin{equation}
    \overline{P} \equiv
    \left[\begin{array}{cc}
         \widetilde{P}& 0 \\
         \delta & 1 
    \end{array}\right],\label{eq:extended_transition_matrix}
\end{equation}
where $\delta$ is defined such that $\sum_j \overline{P}_{ji} = 1$, i.e. $\delta_i = 1 - e^{\beta r_i}$.

\begin{figure}
    \centering
    \includegraphics{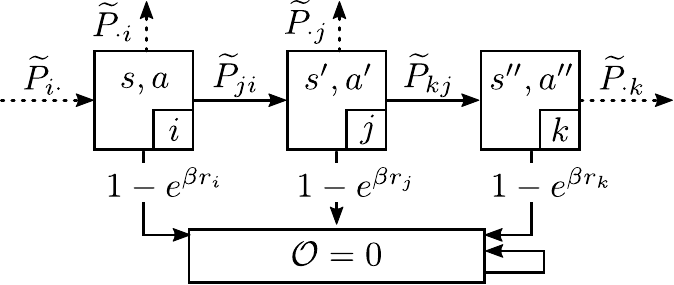}
    \caption{System dynamics in the extended model with transition matrix $\overline{P}$.  Transition $i\to \mathcal{O}=0$ occurs with probability $1-e^{\beta r_{i}}$. The introduction of an absorbing state provides an interpretation for the binary random variable $\mathcal{O}$. Conditioning on optimality (i.e. $\mathcal{O}=1$) is equivalent to conditioning on non-absorption.}
    \label{fig:optimality_diagram}
\end{figure}

The extended model introduced above provides an interpretation for the optimality variable introduced in Eq.~\ref{eq:optimality_var_single_step} as specifying the probability of non-absorption (see Fig.~\ref{fig:optimality_diagram}). Let us consider the system's evolution for $N$ time steps using the transition matrix $\overline{P}$.
Imposing the condition $\mathcal{O}_{1:N}$ is equivalent to conditioning on non-absorption for all $N$ time steps. Thus the optimal trajectory distribution is generated by considering  the probability distribution over trajectories generated by $\overline{P}$, conditional on no transitions to the absorbing state for the entire trajectory. This interpretation allows us to make connections to the theory of quasi-stationary distributions~\cite{meleard2012quasi,van2013quasi} which can be used to analyze Markovian processes conditioned on non-absorption.  

For the dynamics generated by $\overline{P}$, given an initial state-action pair $i$, the probability of transitioning to state-action pair $j$ after taking $N$ steps is given by  $\left[ \widetilde{P}^N \right]_{ji}$. 
In the following, we assume that
$\widetilde{P}$ is a primitive matrix, meaning that the corresponding dynamics is irreducible and aperiodic. 
In this case, the Perron-Frobenius theorem
implies that $\widetilde{P}$ has a unique dominant eigenvalue $\rho$ with a corresponding unique right eigenvector $\mathbf{v}$ (with $v_i > 0$) and a unique left eigenvector $\mathbf{u}$ (with $u_i > 0$). The normalization of the eigenvectors is chosen such that $\sum_i v_i = 1$ and $\sum_i u_iv_i = 1$~\cite{meleard2012quasi}. Furthermore since $\widetilde{P}$ is sub-stochastic (column sums between $0$ and $1$), we must have $\rho < 1$ and so we define $\theta > 0$ such that $\rho = e^{-\beta\theta}$.

We now consider the limit of large $N$, for which, using the spectral decomposition of $\widetilde{P}$, we have
\begin{equation}
    \left[ \widetilde{P}^N \right]_{ji} \approx e^{-\beta\theta N} u_iv_j
\end{equation}
 Furthermore, let $e^{-\beta \xi}$ denote the magnitude of the next dominant eigenvalue. Then the convergence of the preceding equation is exponential in $N$, i.e. the condition determining the long-time limit corresponds to $ e^{-N \beta(\xi - \theta)} \ll 1$.
 
Now the probability that a trajectory starting with state-action pair $z_1=(s_{1},a_{1})$ is optimal for $N$ steps is given by
\begin{equation}
    P(\mathcal{O}_{1:N}|z_1=i) = \sum_j \left[ \widetilde{P}^N \right]_{ji}  \approx e^{-\beta\theta N}u_i \label{eq:optimality_eqn}
\end{equation}
This result can be used to derive the posterior distribution over trajectories conditioned on optimality. 
Typically, the difficulty in deriving expressions for the posterior distribution stems from estimating the partition sum in the denominator of Eq.~\ref{eq:bayes_rhs}. However, we note that the partition sum is given by $P(\mathcal{O}_{1:N}) = \sum_{i} p({z=i}) P(\mathcal{O}_{1:N}|{z=i})$ and thus can be estimated using the results derived. 

To derive expressions for the posterior dynamics and state distributions conditioned on optimality, we define, consistent with the terminology in NESM, the \textit{driven transition matrix}
\begin{equation}
    \left[P_d\right]_{ji} = p\left(z'=j|z=i,\mathcal{O}_{1:N}\right).
    \label{eq:driven_transition_matrix}
\end{equation}
This definition implies that the driven transition matrix is the generator of the 
Markov chain corresponding to the optimal dynamics.
In the long-time limit, we obtain that the driven matrix is given by (see Appendix~\ref{appendix:optimal_distributions}, section~\ref{app_sec:driven_dynamics})
\begin{equation}
   [P_d]_{ji} = \frac{\widetilde{P}_{ji} u_j}{e^{-\beta\theta} u_i},
   \label{eq:Doob-h-transform}
\end{equation}
which recovers the expression for the driven model as a generalized Doob $h$-transform in recent work in NESM \cite{jack2010large,chetrite2015nonequilibrium,chetrite2015variational}. It is interesting to note that our analysis recovers this result based on Bayesian inference of the posterior trajectory distribution.

The result for the driven matrix can be used to derive the following expressions for the optimal dynamics, policy and initial state-action pair distributions (see Appendix~\ref{appendix:optimal_distributions}, sections~\ref{app_sec:optimal_policy} and~\ref{app_sec:optimal_transition_dynamics})
\begin{align}
    p(s'|s,a, \mathcal{O}_{1:N})
    &= \frac{p(s'|s,a)e^{\beta r(s,a)}}{e^{-\beta\theta}u(s,a)}\sum_{a'}u(s',a')\pi(a'|s')\\ 
    \pi(a|s, \mathcal{O}_{1:N}) &= \frac{u(s,a)\pi(a|s)}{\sum_{a'}u(s,a')\pi(a'|s)}
    \\ 
    p(s_{1},a_{1}| \mathcal{O}_{1:N}) &= \frac{p(s_{1},a_{1}) u(s_{1},a_{1}) }{\sum_{(s'_1,a'_1)} p(s'_{1},a'_{1}) u(s'_{1},a'_{1})}
     \label{eq:optimal_dynamics_equations}
\end{align}

The preceding equations, one of the main results of this paper, show that in the long-time limit the optimal dynamics can be completely characterized by the dominant eigenvalue and the corresponding left eigenvector of the tilted matrix $\widetilde{P}$. While previous work has shown how a special class of MDPs are linearly solvable~\cite{todorov2006,Todorov2009Jul}, our results show that linear solutions can be obtained for more general MDP models in the long-time limit.

The significance of this result is that it provides a closed-form solution for the central problem of entropy-regularized RL (stated in Eq.~\ref{eq:bayes_lhs}). For the case of deterministic dynamics, the results show that the optimal dynamics is unchanged from the original dynamics and the optimal policy is determined by the left eigenvector $\mathbf{u}$. For the case of stochastic dynamics, the results allow us to determine how the original dynamics must be controlled to obtain the optimal dynamics.

\begin{figure}
    \centering
    \def\svgwidth{3.4in}
%% Creator: Inkscape 1.1 (c68e22c387, 2021-05-23), www.inkscape.org
%% PDF/EPS/PS + LaTeX output extension by Johan Engelen, 2010
%% Accompanies image file 'fig2.eps' (pdf, eps, ps)
%%
%% To include the image in your LaTeX document, write
%%   \input{<filename>.pdf_tex}
%%  instead of
%%   \includegraphics{<filename>.pdf}
%% To scale the image, write
%%   \def\svgwidth{<desired width>}
%%   \input{<filename>.pdf_tex}
%%  instead of
%%   \includegraphics[width=<desired width>]{<filename>.pdf}
%%
%% Images with a different path to the parent latex file can
%% be accessed with the `import' package (which may need to be
%% installed) using
%%   \usepackage{import}
%% in the preamble, and then including the image with
%%   \import{<path to file>}{<filename>.pdf_tex}
%% Alternatively, one can specify
%%   \graphicspath{{<path to file>/}}
%% 
%% For more information, please see info/svg-inkscape on CTAN:
%%   http://tug.ctan.org/tex-archive/info/svg-inkscape
%%
\begingroup%
  \makeatletter%
  \providecommand\color[2][]{%
    \errmessage{(Inkscape) Color is used for the text in Inkscape, but the package 'color.sty' is not loaded}%
    \renewcommand\color[2][]{}%
  }%
  \providecommand\transparent[1]{%
    \errmessage{(Inkscape) Transparency is used (non-zero) for the text in Inkscape, but the package 'transparent.sty' is not loaded}%
    \renewcommand\transparent[1]{}%
  }%
  \providecommand\rotatebox[2]{#2}%
  \newcommand*\fsize{\dimexpr\f@size pt\relax}%
  \newcommand*\lineheight[1]{\fontsize{\fsize}{#1\fsize}\selectfont}%
  \ifx\svgwidth\undefined%
    \setlength{\unitlength}{180bp}%
    \ifx\svgscale\undefined%
      \relax%
    \else%
      \setlength{\unitlength}{\unitlength * \real{\svgscale}}%
    \fi%
  \else%
    \setlength{\unitlength}{\svgwidth}%
  \fi%
  \global\let\svgwidth\undefined%
  \global\let\svgscale\undefined%
  \makeatother%
  \begin{picture}(1,0.39999999)%
    \lineheight{1}%
    \setlength\tabcolsep{0pt}%
    \put(0,0){\includegraphics[width=\unitlength]{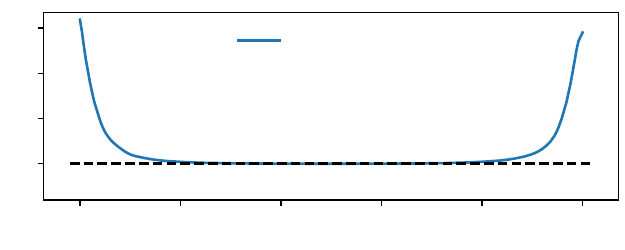}}%
    \put(0.11907008,0.03946189){\color[rgb]{0,0,0}\makebox(0,0)[lt]{\lineheight{1.25}\smash{\begin{tabular}[t]{l}0\end{tabular}}}}%
    \put(0.27107645,0.03946189){\color[rgb]{0,0,0}\makebox(0,0)[lt]{\lineheight{1.25}\smash{\begin{tabular}[t]{l}50\end{tabular}}}}%
    \put(0.42308331,0.03946189){\color[rgb]{0,0,0}\makebox(0,0)[lt]{\lineheight{1.25}\smash{\begin{tabular}[t]{l}100\end{tabular}}}}%
    \put(0.58392224,0.03946189){\color[rgb]{0,0,0}\makebox(0,0)[lt]{\lineheight{1.25}\smash{\begin{tabular}[t]{l}150\end{tabular}}}}%
    \put(0.74476117,0.03946189){\color[rgb]{0,0,0}\makebox(0,0)[lt]{\lineheight{1.25}\smash{\begin{tabular}[t]{l}200\end{tabular}}}}%
    \put(0.9056001,0.03946189){\color[rgb]{0,0,0}\makebox(0,0)[lt]{\lineheight{1.25}\smash{\begin{tabular}[t]{l}250\end{tabular}}}}%
    \put(0.45444444,0.0000000){\color[rgb]{0,0,0}\makebox(0,0)[lt]{\lineheight{1.25}\smash{\begin{tabular}[t]{l}Time step $t$\end{tabular}}}}%
    \put(0.00637155,0.12741523){\color[rgb]{0,0,0}\makebox(0,0)[lt]{\lineheight{1.25}\smash{\begin{tabular}[t]{l}0.0\end{tabular}}}}%
    \put(0.00637155,0.19986865){\color[rgb]{0,0,0}\makebox(0,0)[lt]{\lineheight{1.25}\smash{\begin{tabular}[t]{l}2.5\end{tabular}}}}%
    \put(0.00637155,0.27232194){\color[rgb]{0,0,0}\makebox(0,0)[lt]{\lineheight{1.25}\smash{\begin{tabular}[t]{l}5.0\end{tabular}}}}%
    \put(0.00637155,0.344775){\color[rgb]{0,0,0}\makebox(0,0)[lt]{\lineheight{1.25}\smash{\begin{tabular}[t]{l}7.5\end{tabular}}}}%
    \put(0.47399744,0.32608611){\color[rgb]{0,0,0}\makebox(0,0)[lt]{\lineheight{1.25}\smash{\begin{tabular}[t]{l}$D_{\text{KL}}(t)$\end{tabular}}}}%
    \put(0.47399744,0.27465269){\color[rgb]{0,0,0}\makebox(0,0)[lt]{\lineheight{1.25}\smash{\begin{tabular}[t]{l}\end{tabular}}}}%
  \end{picture}%
\endgroup%
    \caption{
    Comparison of the optimal state-action pair distribution and its approximation using the Perron eigenvectors of the tilted matrix $\widetilde{P}$, as a function of time step $t$, with $N=250$. The Kullback–Leibler divergence between the exact, time-dependent distribution and the bulk/stationary distribution estimated using Eq.~\ref{eq:steady_state_distribution} is shown. The plot shows that the ratio $\frac{u(s_t,a_t)v(s_t,a_t)}{p(s_t,a_t | \mathcal{O}_{1:T} )} \approx 1$ in the ``bulk" region of the trajectory. 
    }
    \label{fig:stationary_distribution_approximation}

\end{figure}
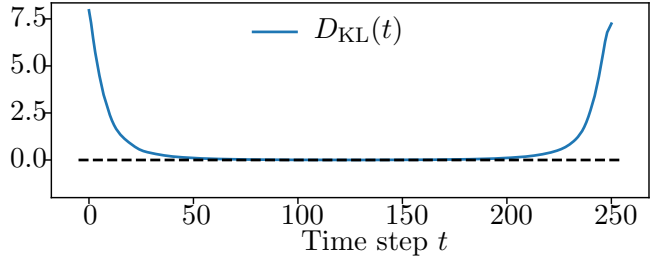

\section{Value functions and statistical mechanics}
The results derived for the optimal dynamics can be used to derive analytical expressions for optimal value functions in entropy-regularized RL (also called soft value functions~\cite{Levine2018May} and denoted by $Q(s,a)$ and $V(s)$) and to make further connections to statistical mechanics.
The optimal value function $Q(s,a)$ represents the expected future return to be collected, given that action $a$ is taken from the initial state $s$, and the optimal dynamics and policy are followed thereafter.
Note that this expected future return includes the penalization given by the entropic cost term $\beta^{-1}\mathcal{H}$ (see Eq.~\ref{eq:objective_function}). 
Specifically $Q(s,a)$ is obtained by maximizing the average return over the controlled trajectory distribution: $\mathbb{E}_{p_c(\tau|s,a)} [R_\tau] - \frac{1}{\beta}\mathcal{H}(p_c(\tau|s,a)||p(\tau|s,a))$. Note that, if we instead consider the energetic costs over trajectories (i.e. $E_\tau = -R_\tau$), the problem of maximizing average returns is equivalent to the problem of minimizing average costs: $\mathbb{E}_{p_c(\tau|s,a)} [E_\tau] + \frac{1}{\beta}\mathcal{H}(p_c(\tau|s,a)||p(\tau|s,a))$, in correspondence with Eq.~\ref{eq:objective_function}.
In the following, we show how this optimization problem can be solved by connecting to the free energy concept from statistical mechanics.

To find the optimal value function, we need to consider the trajectory distribution corresponding to optimal control. Conditioned on the first step $z_1 = (s, a)$, the optimal trajectory distribution is given by
\begin{equation}
    p(\tau|s,a,\mathcal{O}_{1:N}) = \frac{1}{Z_p(s,a)}p(\tau|s,a)e^{-\beta E_\tau},\label{eq:boltzmann_distribution}
\end{equation}
where $Z_p(s,a)=\sum_\tau {p(\tau|s,a)} e^{-\beta E_\tau}$ can be regarded as the partition function corresponding to the non-equilibrium free energy function
\begin{equation}
    F_p(s,a) = - \frac{1}{\beta} \log Z_p(s,a).
\end{equation}

We note that the free energy defined above corresponds to the lower bound of the entropy-regularized RL objective, representing the minimized expected total cost with both energetic and entropic contributions \cite{theodorou2015}
\begin{equation*}
    F_p(s,a) \leq \mathbb{E}_{p_c(\tau|s,a)} [E_\tau] + \frac{1}{\beta}\mathcal{H}(p_c(\tau|s,a)||p(\tau|s,a)),
\end{equation*}
and equality is attained when the controlled trajectory distribution is given by  Eq.~\ref{eq:boltzmann_distribution}. Thus the problem of minimizing the expected costs, or equivalently maximizing the expected return, is solved by the free energy, and correspondingly, we obtain $Q(s,a)=-F_p(s,a)$. In other words, the function that maximizes the expected total returns in entropy-regularized RL ($Q(s,a)$) is given by $e^{\beta Q(s,a)} = Z_p(s,a) = p(\mathcal{O}_{1:N}|s,a)$, consistent with \cite{Levine2018May}.
This result,
in combination with Eq.~\ref{eq:optimality_eqn} and the definition of the state-dependent value function $V(s)$: $e^{\beta V(s)} = \sum_a \pi(a|s)e^{\beta Q(s,a)}$, yields the relations
\begin{align}
    \beta Q(s,a) &= -\beta\theta N + \log u(s,a)\\
    \beta V(s) &= -\beta\theta N + \log \sum_{a} \pi(a|s)u(s,a).
    \label{eq:value_functions}
\end{align}
Thus the value functions in entropy-regularized RL can be obtained using the dominant eigenvalue and the left Perron eigenvector of the tilted matrix $\widetilde{P}$. These results have been validated by comparing with the dynamic programming solution for entropy-regularized RL (see Appendix~\ref{appendix:experimental_validation}). The significance of the preceding equations is that they provide a mapping between problems of interest in NESM and entropy-regularized RL such that approaches from one field can be used to solve problems in the other. For example, using the derived equations, function approximators, a popular tool in deep reinforcement learning for estimating value functions \cite{mnih2013playing}, can potentially be used as a new method for calculating the left and right dominant eigenvectors of the tilted generator in NESM.

Besides  the  value  functions,  other  quantities  of  interest in RL can also be obtained using the Perron-Frobenius eigenvalue  and the corresponding  eigenvectors, as previously noted in diverse systems of interest~\cite{giardina,jack2010large,meleard2012quasi,nemoto}.
For example, in the long-time limit the right eigenvector
gives the probability of observing a state-action pair conditioned on optimality: $p(s_{t}, a_{t}|\mathcal{O}_{1:t-1}) = v(s_{t},a_{t})$.
Using Eq.~\ref{eq:optimal_dynamics_equations}, for $t$ such that $t \to \infty$ and $(N-t) \to \infty$ (i.e., the ``bulk" region of the trajectory), we also have (see Appendix~\ref{appendix:optimal_distributions}, section~\ref{app_sec:optimal_steady_state_distribution})
\begin{equation}
    p(s_{t},a_{t}|\mathcal{O}_{1:N}) \approx u(s_{t},a_{t})v(s_{t},a_{t}).
    \label{eq:steady_state_distribution}
\end{equation}
We note that $u(s,a) v(s,a)$ represents the components of the dominant right eigenvector of the driven matrix $P_d$, i.e. the components of the steady-state distribution over state-action pairs generated by the driven dynamics. 

As shown in Fig.~\ref{fig:stationary_distribution_approximation}, the exact optimal state-action pair distribution is in excellent agreement with the approximation obtained using the steady-state distribution of the driven dynamics, for time $t$ in the ``bulk'' region of the trajectory (i.e. far from the extremities at $t=0$ and $t=N$). Given that the steady-state distribution over state-action pairs is a quantity of significant interest in RL applications such as Inverse RL~\cite{fu2018learning}, the result obtained in Eq.~\ref{eq:steady_state_distribution} can significantly impact the computations involved in such RL approaches. 

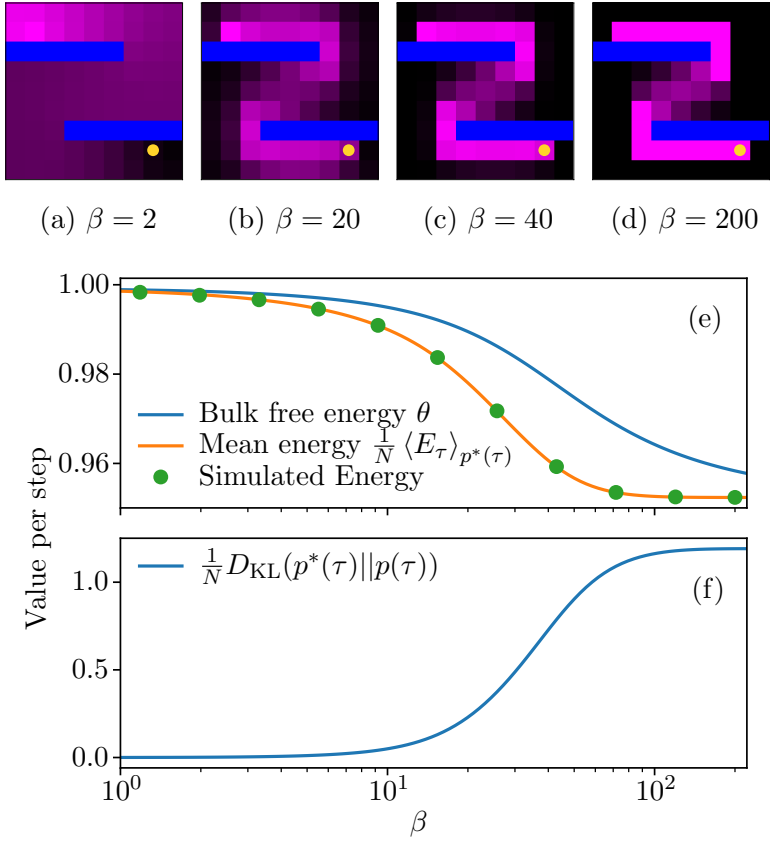
\begin{figure}
    \def\svgwidth{4in}
%% Creator: Inkscape 1.1 (c68e22c387, 2021-05-23), www.inkscape.org
%% PDF/EPS/PS + LaTeX output extension by Johan Engelen, 2010
%% Accompanies image file 'fig3_panel.eps' (pdf, eps, ps)
%%
%% To include the image in your LaTeX document, write
%%   \input{<filename>.pdf_tex}
%%  instead of
%%   \includegraphics{<filename>.pdf}
%% To scale the image, write
%%   \def\svgwidth{<desired width>}
%%   \input{<filename>.pdf_tex}
%%  instead of
%%   \includegraphics[width=<desired width>]{<filename>.pdf}
%%
%% Images with a different path to the parent latex file can
%% be accessed with the `import' package (which may need to be
%% installed) using
%%   \usepackage{import}
%% in the preamble, and then including the image with
%%   \import{<path to file>}{<filename>.pdf_tex}
%% Alternatively, one can specify
%%   \graphicspath{{<path to file>/}}
%% 
%% For more information, please see info/svg-inkscape on CTAN:
%%   http://tug.ctan.org/tex-archive/info/svg-inkscape
%%
\begingroup%
  \makeatletter%
  \providecommand\color[2][]{%
    \errmessage{(Inkscape) Color is used for the text in Inkscape, but the package 'color.sty' is not loaded}%
    \renewcommand\color[2][]{}%
  }%
  \providecommand\transparent[1]{%
    \errmessage{(Inkscape) Transparency is used (non-zero) for the text in Inkscape, but the package 'transparent.sty' is not loaded}%
    \renewcommand\transparent[1]{}%
  }%
  \providecommand\rotatebox[2]{#2}%
  \newcommand*\fsize{\dimexpr\f@size pt\relax}%
  \newcommand*\lineheight[1]{\fontsize{\fsize}{#1\fsize}\selectfont}%
  \ifx\svgwidth\undefined%
    \setlength{\unitlength}{252.00000433bp}%
    \ifx\svgscale\undefined%
      \relax%
    \else%
      \setlength{\unitlength}{\unitlength * \real{\svgscale}}%
    \fi%
  \else%
    \setlength{\unitlength}{\svgwidth}%
  \fi%
  \global\let\svgwidth\undefined%
  \global\let\svgscale\undefined%
  \makeatother%
  \begin{picture}(1,1.09907322)%
    \lineheight{1}%
    \setlength\tabcolsep{0pt}%
    \put(0,0){\includegraphics[width=\unitlength]{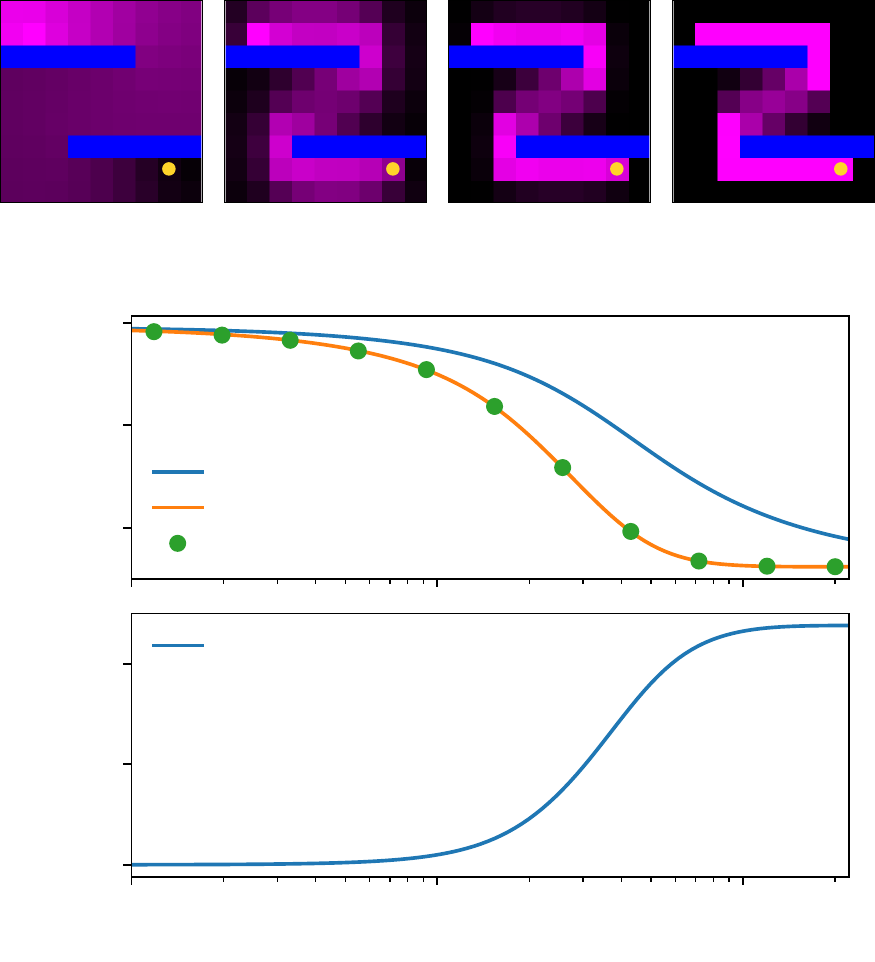}}%
    \put(0.04444040,0.80475682){\color[rgb]{0,0,0}\makebox(0,0)[lt]{\lineheight{1.25}\smash{\begin{tabular}[t]{l}(a) $\beta = 2$\end{tabular}}}}%
    \put(0.28986901,0.80475682){\color[rgb]{0,0,0}\makebox(0,0)[lt]{\lineheight{1.25}\smash{\begin{tabular}[t]{l}(b) $\beta = 20$\end{tabular}}}}%
    \put(0.54583365,0.80475682){\color[rgb]{0,0,0}\makebox(0,0)[lt]{\lineheight{1.25}\smash{\begin{tabular}[t]{l}(c) $\beta = 40$\end{tabular}}}}%
    \put(0.79138446,0.80475682){\color[rgb]{0,0,0}\makebox(0,0)[lt]{\lineheight{1.25}\smash{\begin{tabular}[t]{l}(d) $\beta = 200$\end{tabular}}}}%
    \put(0.06901255,0.48531296){\color[rgb]{0,0,0}\makebox(0,0)[lt]{\lineheight{1.25}\smash{\begin{tabular}[t]{l}0.96\end{tabular}}}}%
    \put(0.06901255,0.60230183){\color[rgb]{0,0,0}\makebox(0,0)[lt]{\lineheight{1.25}\smash{\begin{tabular}[t]{l}0.98\end{tabular}}}}%
    \put(0.06901255,0.71929069){\color[rgb]{0,0,0}\makebox(0,0)[lt]{\lineheight{1.25}\smash{\begin{tabular}[t]{l}1.00\end{tabular}}}}%
    \put(0.25308374,0.54991295){\color[rgb]{0,0,0}\makebox(0,0)[lt]{\lineheight{1.25}\smash{\begin{tabular}[t]{l}Bulk free energy $\theta$\end{tabular}}}}%
    \put(0.25308374,0.50915743){\color[rgb]{0,0,0}\makebox(0,0)[lt]{\lineheight{1.25}\smash{\begin{tabular}[t]{l}Mean energy $\frac{1}{N}\left<E_{\tau}\right>_{p^*(\tau)}$\end{tabular}}}}%
    \put(0.25308374,0.46840462){\color[rgb]{0,0,0}\makebox(0,0)[lt]{\lineheight{1.25}\smash{\begin{tabular}[t]{l}Simulated Energy\end{tabular}}}}%
    \put(0.12530615,0.05566469){\color[rgb]{0,0,0}\makebox(0,0)[lt]{\lineheight{1.25}\smash{\begin{tabular}[t]{l}$10^0$\end{tabular}}}}%
    \put(0.47476940,0.05590781){\color[rgb]{0,0,0}\makebox(0,0)[lt]{\lineheight{1.25}\smash{\begin{tabular}[t]{l}$10^1$\end{tabular}}}}%
    \put(0.82423033,0.05566469){\color[rgb]{0,0,0}\makebox(0,0)[lt]{\lineheight{1.25}\smash{\begin{tabular}[t]{l}$10^2$\end{tabular}}}}%
    \put(0.52883872,0.01497444){\color[rgb]{0,0,0}\makebox(0,0)[lt]{\lineheight{1.25}\smash{\begin{tabular}[t]{l}$\beta$\end{tabular}}}}%
    \put(0.08667755,0.10022189){\color[rgb]{0,0,0}\makebox(0,0)[lt]{\lineheight{1.25}\smash{\begin{tabular}[t]{l}0.0\end{tabular}}}}%
    \put(0.08667755,0.21501356){\color[rgb]{0,0,0}\makebox(0,0)[lt]{\lineheight{1.25}\smash{\begin{tabular}[t]{l}0.5\end{tabular}}}}%
    \put(0.08667755,0.3298047){\color[rgb]{0,0,0}\makebox(0,0)[lt]{\lineheight{1.25}\smash{\begin{tabular}[t]{l}1.0\end{tabular}}}}%
    % \put(0.25308374,0.35181581){\color[rgb]{0,0,0}\makebox(0,0)[lt]{\lineheight{1.25}\smash{\begin{tabular}[t]{l}$\frac{1}{N}D_{kl}(\pi^*(a|s)||\pi(a|s))$\end{tabular}}}}%
    \put(0.25308374,0.35181581){\color[rgb]{0,0,0}\makebox(0,0)[lt]{\lineheight{1.25}\smash{\begin{tabular}[t]{l}$\frac{1}{N}D_{\text{KL}}(p^*(\tau)||p(\tau))$\end{tabular}}}}%
    \put(0.05212884,0.27842507){\color[rgb]{0,0,0}\rotatebox{90}{\makebox(0,0)[lt]{\lineheight{1.25}\smash{\begin{tabular}[t]{l}Value per step\end{tabular}}}}}%
    \put(0.89210259,0.67439965){\color[rgb]{0,0,0}\makebox(0,0)[lt]{\lineheight{1.25}\smash{\begin{tabular}[t]{l}(e)\end{tabular}}}}%
    \put(0.89759112,0.31919444){\color[rgb]{0,0,0}\makebox(0,0)[lt]{\lineheight{1.25}\smash{\begin{tabular}[t]{l}(f)\end{tabular}}}}%
  \end{picture}%
\endgroup%
    \caption{Results for a 9 by 9 maze, trajectory length $N=10^4$. Figures (a) - (d) show how state occupation frequencies (derived from the optimal trajectory distribution) change with temperature. Figures (e) and (f) show the mean energetic costs, and relative entropy per time-step as functions of $\beta$.}
    \label{fig:energy entropy and beta}
\end{figure}

To further validate the theory presented, we consider the ``gridworld'' setting shown in Fig.~\ref{fig:energy entropy and beta}(a)-(d) in which an agent can take actions $a$ by deterministically moving up, down, left, or right. The state $s$ of the agent is simply the grid cell in which it resides. The agent's task is to navigate to the only rewarding state: the goal, indicated by the yellow circle. The initial state of the agent is in the top left part of the maze. The shading of states represents the steady-state distribution $\sum_{a} u(s,a)v(s,a)$ for various values of the control parameter, $\beta$. We note that as $\beta~\to~\infty$, the agent acts greedily by not deviating from the shortest path, that is, the most probable trajectories are those with higher rewards. This observed behaviour reveals the role of the $\beta$ parameter, which is to control the preference of the agent to purely minimize energy (maximize rewards) in exchange for stochasticity.

In the limit of large $N$, we have $ \frac{F(s,a)}{N} \to \theta$, which can be interpreted as the ``bulk" free energy per time step. Furthermore,
we can also obtain approximations for quantities of interest such as the mean energetic cost per time-step, through the steady state distribution in Eq.~\ref{eq:steady_state_distribution}, resulting in the following expression: 
\begin{equation}
    \frac{1}{N}\mathbb{E}[E_\tau] = - \sum_{s,a} u(s,a)v(s,a)r(s,a). \label{eq:e_cost_estimation}\\
\end{equation}
As shown in Fig.~\ref{fig:energy entropy and beta}(e), the preceding equation is in excellent agreement with results from simulations.
We further note that as the inverse temperature parameter $\beta$ is varied, the optimal trajectory distribution switches from primarily minimizing entropic costs at high temperatures (low $\beta$) to primarily minimizing energetic costs at low temperatures (high $\beta$).
The approach developed therefore not only enables us to obtain the value functions of interest in entropy-regularized RL, but to also derive analytical expressions for the energetic and entropic contributions, which were previously unavailable.

\section{u-$\theta$ Learning} 
The framework developed shows how several quantities of interest in entropy-regularized RL can be obtained using the dominant eigenvalue and the corresponding left eigenvector of the tilted matrix. 
In the following, we show how these quantities can be obtained in a {\em model-free} setting (that is, without explicit knowledge of the dynamics and rewards) 
 by allowing the agent to collect experience by randomly exploring using the original transition dynamics.

By taking the sum over the columns of the driven matrix in Eq.~\ref{eq:Doob-h-transform}, we note that the left eigenvector elements can be written as an expectation value over the original transition dynamics. Correspondingly, the dominant eigenvalue and left eigenvector can be obtained through a learning process based on the following equation
\begin{equation}
    u(s, a) e^{-\beta\theta} = e^{\beta r(s, a)}  \mathbb{E}_{\sim p(s',a'|s, a)}[ u(s', a')]
    \label{eq:utheta_backup_eqn}
\end{equation}

The corresponding update equations for learning $u(s,a)$ and $\theta$ are
\begin{align}
u(s, a) &\leftarrow (1-\alpha) u(s,a) + \alpha \frac{ e^{\beta r(s, a)} }{e^{-\beta\theta}} u(s', a')
\label{eq:td_update_1}\\
e^{-\beta\theta} &\leftarrow (1-\alpha_\theta) e^{-\beta\theta} + \alpha_\theta e^{\beta r(s, a)} \frac{ u(s', a')}{u(s, a)}.
\label{eq:td_update_2}
\end{align}
Where $\alpha$ and $\alpha_\theta$ are their respective learning rates \cite{Sutton2018}. 
Further refinements of the algorithm outlined above can be developed following the connections to learning algorithms for risk-sensitive control~\cite{borkar2010learning}.
Note that the prior policy is used for sampling actions during the training process (see Eqn.~\ref{eq:utheta_backup_eqn}). Thus this model-free approach to RL, which we term $u$-$\theta$ learning, is fundamentally an {\it off-policy} approach~\cite{levine2020offline} wherein the optimal policy is obtained via system exploration using the prior policy.
Our simulations (see Appendix~\ref{appendix:experimental_validation}) indicate that optimal policies obtained using this method are in excellent agreement with the corresponding results obtained using dynamic programming \cite{bellman1954theory} on the soft Bellmann backup equation. Appendix~\ref{appendix:derivation_of_soft_bellman_backup_equations} shows how the soft Bellmann backup equation arises from the definition of the tilted matrix $\widetilde{P}$.

In conclusion, we have established a mapping between entropy-regularized RL and recent research in NESM using large deviation theory. 
The results derived include analytical expressions for quantities of interest in RL and lead to a novel learning algorithm for model-free RL.
The results obtained have thus established a new framework for analyzing optimization problems using entropy-regularized RL and generalizations of this approach hold promise for obtaining solutions to a broader range of optimization problems in physics and machine learning.

\begin{acknowledgments}
The authors acknowledge funding support from the NSF through Award DMS-1854350.
\end{acknowledgments}

\appendix

\section{Implementation details}

For the purposes of testing and validation we have developed an implementation of the method using Python, and used the Gym environment framework developed by OpenAI~\cite{Brockman2016Jun}. Since we focus on discrete state-action spaces, we shall work with the FrozenLake Gym environment, which we have modified to meet our needs regarding the transition dynamics and reward structure.

The code's implementation includes model-based and model-free solutions, along with example scripts to use our method. The code is made available as Supplemental Material in this publication, and as a Github repository at:
\href{https://github.com/argearriojas/2023-EntRegRL}{github.com/argearriojas/2023-EntRegRL}.

\section{Driven dynamics and optimal distributions}
\label{appendix:optimal_distributions}

\subsection{Driven dynamics}
\label{app_sec:driven_dynamics}

The probability distribution for trajectories, $\tau_{1:T}=(z_1, z_2, \dots, z_T)$ with $z_t=(s_t, a_t)$, conditioned on optimality 
is 
given by (see Eq.~\ref{eq:bayes_rhs})
\begin{equation*}
    p(\tau_{1:T}|\mathcal{O}_{1:T}) 
    = \frac{ p(\tau_{1:T}, \mathcal{O}_{1:T})}{p(\mathcal{O}_{1:T})}
    = \frac{p(\tau) e^{-\beta E_\tau}}{\sum_\tau p(\tau) e^{-\beta E_\tau}},
\end{equation*}
For notational convenience, let $z_t =(s_{t},a_{t})=i$ and $z_{t+1}=(s'_{t+1},a'_{t+1})=j$ denote two 
consecutive state-action tuples in the trajectory $\tau_{1:T}$, with $1 \le t < T$.
The corresponding elements of the driven and tilted matrices are, by definition, 
\begin{align*}
    [P_d]_{ji} &= p(s'_{t+1},a'_{t+1}|s_{t},a_{t},\mathcal{O}_{1:T})
    \\
    [\widetilde{P}]_{ji} &= p(s'_{t+1},a'_{t+1}|s_{t},a_{t})e^{\beta r(s_{t},a_{t})}
    ,
\end{align*}
 From the above equations, it can be seen that the tilted matrix is time-independent whereas the driven matrix will, in general, depend on the time index $t$. In the following, we consider the long-time limit $(T-t) \to \infty$. In this case, we will see that the driven matrix is independent of the time index $t$. \\

Let us divide the trajectory $\tau_{1:T}$ into two parts such that $\tau_{1:t-1}=(z_1, z_2, \dots, z_{t-1})$ and $\tau_{t:T}=(z_t, z_{t+1}, \dots, z_T)$.
We will first focus on $\tau_{t:T}$ in the limit $(T-t) = N \to \infty$.
Using the definition of the driven matrix, we have
\begin{equation}
\begin{split}
p(\tau_{t+2:T},z_{t+1}=j&|z_{t}=i,\mathcal{O}_{t:T}) =\\
&p(\tau_{t+2:T}|z_{t+1}=j,\mathcal{O}_{t+1:T})[P_d]_{ji}
\end{split}
\label{eq:driven_matrix_trajectory}
\end{equation}
Using Eq.~\ref{eq:bayes_rhs}, the LHS of Eq.~\ref{eq:driven_matrix_trajectory} can also be expressed as
\begin{equation}
\begin{split}
p(\tau_{t+2:T},z_{t+1}=j&|z_{t}=i,\mathcal{O}_{t:T}) =\\
&\frac{p(\tau_{t+2:T},\mathcal{O}_{t+1:T}|z_{t+1}=j)}{p(\mathcal{O}_{t:T}|z_{t}=i)} [\widetilde{P}]_{ji}.
\end{split}
\label{eq:tilted_matrix_trajectory}
\end{equation}
In Eq.~\ref{eq:driven_matrix_trajectory} using the substitution
\begin{equation*}
    {p(\tau_{t+2:T}|z_{t+1}=j,\mathcal{O}_{t+1:T})} =\frac{p(\tau_{t+2:T},\mathcal{O}_{t+1:T}|z_{t+1}=j)}{p(\mathcal{O}_{t+1:T}|z_{t+1}=j)},
\end{equation*}
and comparing with Eq.~\ref{eq:tilted_matrix_trajectory}, we get:
\begin{equation}
   [P_d]_{ji} = \frac{[\widetilde{P}]_{ji}~p(\mathcal{O}_{t+1:T}|z_{t+1}=j)}{p(\mathcal{O}_{t:T}|z_{t}=i)}
     \label{eq:driven_tilted_relation_1}
\end{equation}

Taking the long-time limit and approximating the tilted transition matrix using the dominant contribution,
\begin{align}
    P(\mathcal{O}_{t:T}|z_t=i) &= \sum_j \left[ \widetilde{P}^N \right]_{ji}  = e^{-\beta \theta N}u_i
    \label{eq:long_time_optimality}
    \\
    P(\mathcal{O}_{t+1:T}|z_{t+1}=j) &= \sum_k \left[ \widetilde{P}^{N-1} \right]_{kj}  = e^{-\beta \theta (N-1)}u_j .
    \nonumber
\end{align}
Substituting in Eq.~\ref{eq:driven_tilted_relation_1} we find that the driven matrix is given by the Doob $h$-transform (see Eq.~\ref{eq:Doob-h-transform}):
\begin{equation*}
   [P_d]_{ji} = \frac{\widetilde{P}_{ji} ~u_j}{e^{-\beta \theta} u_i}.
\end{equation*}

\subsection{Optimal policy}
\label{app_sec:optimal_policy}

To derive the optimal policy, we begin with the observation
\begin{equation}
 p(s_{t},a_{t}|\mathcal{O}_{t:T}) = \frac{p(s_{t},a_{t})p(\mathcal{O}_{t:T}|s_{t},a_{t})}{\sum_{(s_{t},a_{t})} p(s_{t},a_{t}) p(\mathcal{O}_{t:T}|s_{t},a_{t})}
    \label{eq:optimal_distribution_1}   
\end{equation}
Using the approximation in Eq.~\ref{eq:long_time_optimality}, we can rewrite Eq.~\ref{eq:optimal_distribution_1} as
\begin{equation}
    p(s_{t},a_{t}|\mathcal{O}_{t:T}) = \frac{p(s_{t},a_{t})u(s_{t},a_{t})}{\sum_{s_{t},a_{t}} p(s_{t},a_{t})u(s_{t},a_{t}) }
     \label{eq:state-action distribution}
\end{equation}
Note that the preceding equation is valid for times $t$ such that $(T-t) \gg 1$. In particular, it can be applied for the initial time-step to obtain the optimal initial state-action pair distribution result derived in the main text. 
For general $t$, the optimal state distribution can be obtained from Eq.~\ref{eq:state-action distribution} as
\begin{equation*}
    p(s_{t}|\mathcal{O}_{t:T}) = \frac{\sum_{a_{t}} p(s_{t},a_{t})u(s_{t},a_{t})}{\sum_{s_{t},a_{t}} p(s_{t},a_{t})u(s_{t},a_{t}) }.
\end{equation*}
From the preceding equations, we see that, in the long-time limit $(T-t) \to \infty$, the optimal state-action pair distribution is time-independent.
Therefore, using these equations and suppressing the time index, we obtain that the optimal policy is given by
\begin{align}
    p(a|s,\mathcal{O}_{1:T})
    &=
    \frac
    {p(a|s)u(s,a)}
    {\sum_{a} p(a|s)u(s,a) }\nonumber\\
    \pi^*(a|s)
    &=
    \frac
    {\pi(a|s)u(s,a)}
    {\sum_{a} \pi(a|s)u(s,a) },\label{eq:optimal_policy}
\end{align}
where $\pi^*(a|s)$ denotes the optimal policy and  $\pi(a|s)$ is the prior policy.

\subsection{Optimal transition dynamics}
\label{app_sec:optimal_transition_dynamics}

To derive the optimal transition dynamics, we first write Eq.~\ref{eq:Doob-h-transform} as
\begin{align}
    p(s',a'|s,a,\mathcal{O}_{1:T})
    &=
    \frac
    {p(s',a'|s,a)e^{\beta r(s,a)}u(s',a')}
    {e^{-\beta \theta}u(s,a)}\nonumber
    \\
    \pi^*(a'|s')p^*(s'|s,a)
    &=
    \frac
    {\pi(a'|s')p(s'|s,a)e^{\beta r(s,a)}u(s',a')}
    {e^{-\beta \theta}u(s,a)}.
    \label{eq:factorized_h-transform}
\end{align}
By substituting the optimal policy in Eq.~\ref{eq:optimal_policy} into Eq.~\ref{eq:factorized_h-transform}, we find that the optimal transition dynamics is given by
\begin{equation}
    p^*(s'|s,a)
    =
    \frac
    {p(s'|s,a)e^{\beta r(s,a)}}
    {e^{-\beta \theta}u(s,a)}
    \sum_{a'} \pi(a'|s')u(s',a')
    .
    \nonumber
\end{equation}

\subsection{Optimal steady-state distribution}
\label{app_sec:optimal_steady_state_distribution}

Now we consider the initial part of the trajectory $\tau_{1:t-1}$. Consider
\begin{equation}
p(z_{t}=j|z_{1}=i,\mathcal{O}_{1:t-1})  = \frac{p(z_{t}=j,\mathcal{O}_{1:t-1}|z_{1}=i)}{p(\mathcal{O}_{1:t-1}|z_{1}=i)} 
\nonumber
\end{equation}
In the limit $t \to \infty$, using the Perron-Frobenius theorem and Eq.~\ref{eq:long_time_optimality}, we get
\begin{equation}
p(z_{t}=j|z_{1}=i,\mathcal{O}_{1:t-1})  = \frac{e^{-\beta \theta N}~u_{i}v_{j}}{e^{-\beta \theta N}u_i} = v_{j}.
\nonumber
\end{equation}
Thus, the optimal state-action pair distribution at time $t$ is time-independent and independent of the initial state-action pair distribution. This distribution is given by the right eigenvector of the tilted matrix, and is referred to as the quasi-stationary distribution~\cite{meleard2012quasi}.

The preceding equations have shown the equality \mbox{$p(z_{t}=j|\mathcal{O}_{1:t-1}) = v_{j}$}. To obtain the steady-state distribution of the optimal dynamics, we need to derive an expression for $p(z_{t}=j|\mathcal{O}_{1:T})$. To proceed, we split the trajectory in a similar way as above.
\begin{equation}
\begin{split}
p(z_{t}=j&|\mathcal{O}_{1:t-1},\mathcal{O}_{t:T}) = \\
&\frac{p(\mathcal{O}_{t:T}|z_{t}=j,\mathcal{O}_{1:t-1}) p(z_{t}=j|\mathcal{O}_{1:t-1})}
{\sum_{k} p(\mathcal{O}_{t:T}|z_{t}=k,\mathcal{O}_{1:t-1}) p(z_{t}=k|\mathcal{O}_{1:t-1}) } 
\end{split}
\nonumber
\end{equation}
Furthermore, using
\begin{equation}
p(\mathcal{O}_{t:T}|z_{t}=j,\mathcal{O}_{1:t-1}) = p(\mathcal{O}_{t:T}|z_{t}=j) = e^{-\beta \theta N}u_i
\nonumber
\end{equation}
in combination with $p(z_{t}=j|\mathcal{O}_{1:t-1})= v_{j}$, we get
\begin{equation}
p(z_{t}=j|\mathcal{O}_{1:T}) = \frac{e^{-\beta \theta N}u_{j}v_{j}}{e^{-\beta \theta N}\sum_{k} u_{k}v_{k}}  = u_{j}v_{j}
\nonumber
\end{equation}
Thus the optimal state-action pair distribution in the ``bulk'' region of the trajectory (i.e. the times $t$ such that $t \to \infty$ and $(T-t) \to \infty$) is time-independent and is given by the Hadamard product of the left and right eigenvectors of the tilted matrix. It is readily verified that this distribution also corresponds to the steady-state distribution of the driven matrix $P_D$.

\begin{figure*}
    \centering
    \includegraphics[width=0.33\linewidth]{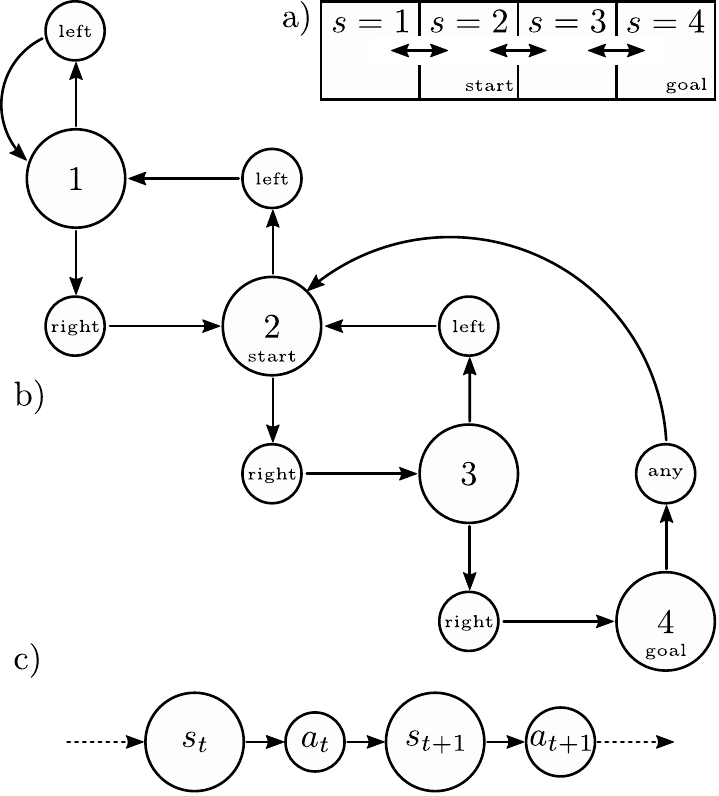}
    \includegraphics[width=0.38\linewidth]{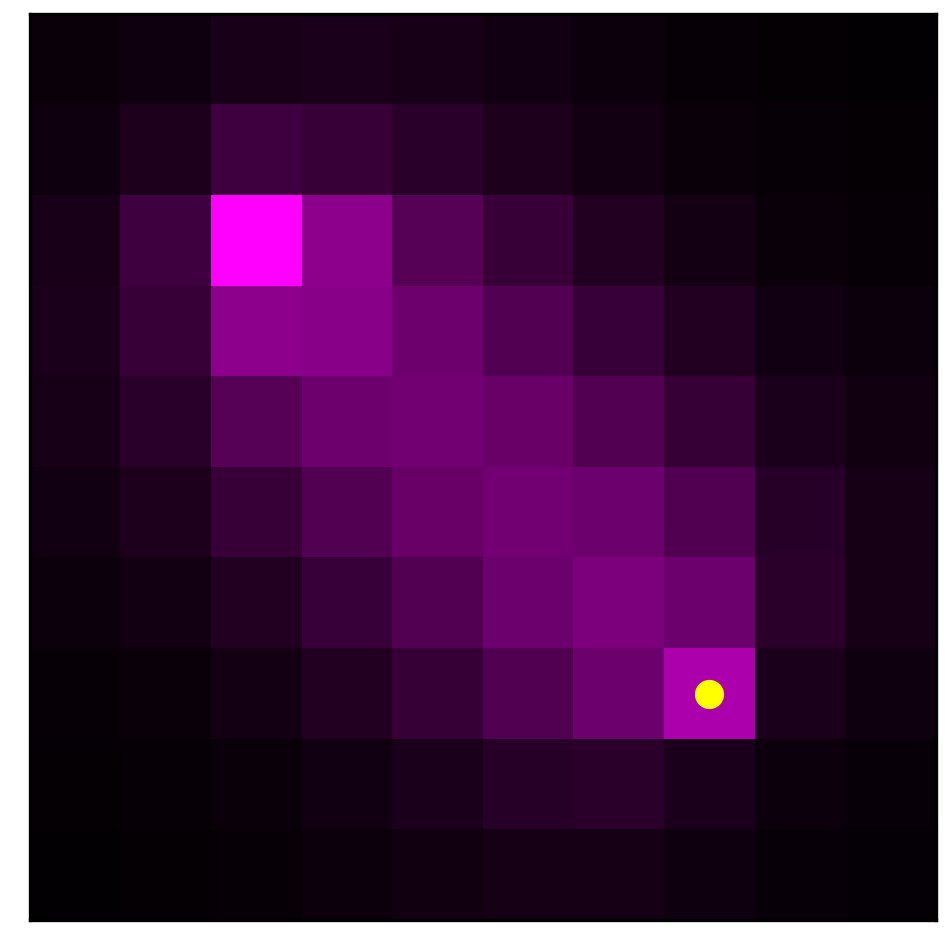}
    \caption{a) A 4 state 1-D maze with two actions available to navigate. b) The corresponding graphical model. Here we model an irreducible Markov chain by making the agent return to the initial state after reaching the goal state. c) Part of the Markov chain at time step $t$. On the right, representation of the stationary distribution resulting from the optimal dynamics, for a larger 10 by 10 2D maze and 4 available actions: left, down, right, up.}
    \label{fig:my_label}
\end{figure*}

\section{Derivation of soft Bellman backup equations}
\label{appendix:derivation_of_soft_bellman_backup_equations}

Recall that we write the indices $i=(s,a)$ and $j=(s', a')$ for two consecutive steps, and the transition matrix is
\begin{equation}
    P_{ji} = p(s',a'|s,a) = p(s'|s,a)\pi(a'|s').
    \nonumber
\end{equation}
From Markov chain theory, when given a transition matrix $P$, we interpret $[P^N]_{ji}$ as the probability of arriving at $j$ after $N$ steps, given that we start from $i$. Since the transition matrix $P$ is a stochastic matrix, we have that $\sum_j [P^N]_{ji}~=~1$. For large $N$, $P^N$ leads to the stationary distribution for the corresponding Markov process.

Let us now consider the tilted transition matrix
\begin{equation}
    \widetilde{P}_{ji} = e^{\beta r_i} P_{ji},
    \nonumber
\end{equation}
which represents a sub-stochastic transition matrix.
As pointed out in the main text, we can expand the graphical model with an extra state in such a way that we obtain a proper stochastic transition matrix. This extra state is an absorbing state, and any trajectory that reaches it is regarded as suboptimal. 
\newline
We can write the probability of remaining optimal after taking $N$ steps in the Markov chain as the probability of non-absorption,
\begin{equation}
    p(\mathcal{O}_{1:N}|s,a) = \sum_j \left[\widetilde{P}^N\right]_{ji}.
    \nonumber
\end{equation}
The preceding equation represents the so-called backward messages \cite{Levine2018May}. Using this we can write a recursive relation which then leads to the soft Bellman backup equation
\begin{align}
    p(\mathcal{O}_{1:N}|s,a) 
    &= \sum_j \sum_m \left[\widetilde{P}^{N-1}\right]_{jm} e^{\beta r_i}P_{mi}
    \nonumber\\
    &= e^{\beta r(s,a)} \sum_{s',a'} p(s',a'|s,a)p(\mathcal{O}_{2:N}|s',a').
    \nonumber
\end{align}

Now, using the definitions of the soft value functions in entropy-regularized RL \cite{Levine2018May}
\begin{align}
    \beta Q(s,a) =& \log p(\mathcal{O}_{1:N}|s,a)
    \nonumber\\
    \beta V(s) =& \log \sum_{a} \pi(a|s) \exp(\beta Q(s,a)),
    \nonumber
\end{align}
we obtain, consistent with the result derived in \cite{Levine2018May}, the following soft backup equation
\begin{align}
    Q(s,a)
    &= r(s,a) + \frac{1}{\beta} \log  \mathbb{E} \left[ \exp(\beta Q(s',a')) \right]
    \nonumber\\
    &= r(s,a) + \frac{1}{\beta} \log \left[ \sum_{s'} p(s'|s,a)  \exp(\beta V(s')) \right],
    \nonumber
\end{align}
where the expectation is taken with respect to the uncontrolled dynamics: the prior policy and the original transition dynamics.

\section{Experimental validation}
\label{appendix:experimental_validation}

In order to validate the analytical framework proposed in the main text and derived here, we defined a series of grid-world mazes for which a complete dynamics model is available, i.e. all available states, actions and transition dynamics are known beforehand. We modified the OpenAI Gym environment ``FrozenLakeEnv'' \cite{Brockman2016Jun}, which has a discrete state-action space. Our version of this environment provides control over the reward function, stochastic behaviour, and an option to define a cyclic mode that results in irreducible MDPs (see Fig.~\ref{fig:my_label}).

\begin{figure}
    \centering
    \includegraphics[width=0.7\linewidth]{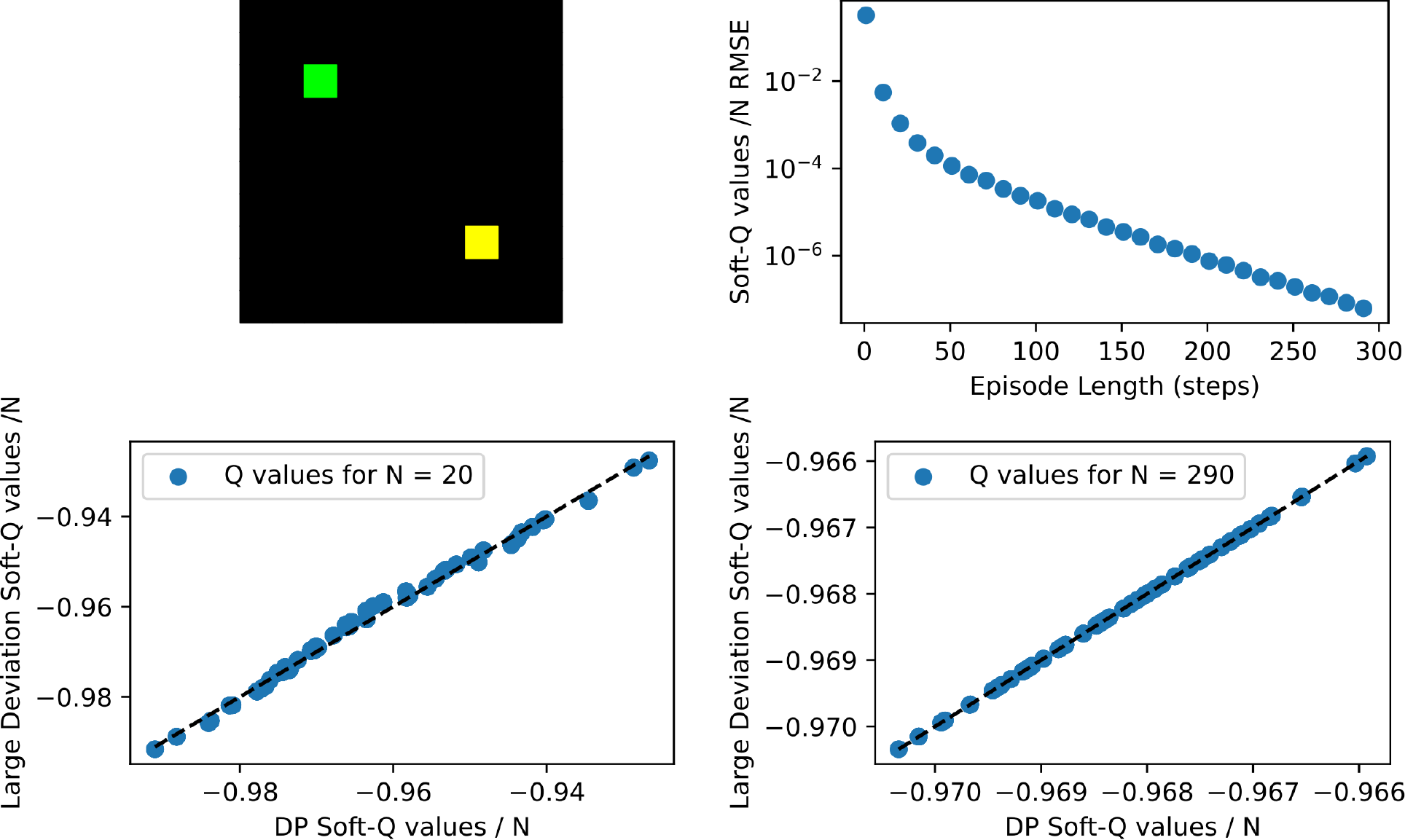}
    \caption{Comparison of the soft-$Q$ values computed by the large deviation approach vs. the dynamic programming solution. Top left: The 10 by 10 empty maze used for the plots in this figure. Top right: Root mean squared deviations of $Q$ values between the large deviation and dynamic programming solutions, as a function of trajectory length. Bottom left: 20 step trajectories. Bottom right: 290 step trajectories. Here we can see perfect correlation between both solutions, for long enough trajectories.}
    \label{fig:ld_vs_dp}
\end{figure}
With this setup, we are able to compute the optimal solution for the objective function in entropy-regularized RL. The resulting soft-$Q$ value function has been compared with the dynamic programming result, which is obtained by directly computing the soft-$Q$ and soft-$V$ value functions at every step (see Fig.~\ref{fig:ld_vs_dp}).
\begin{figure}
    \centering
    \includegraphics[width=0.7\linewidth]{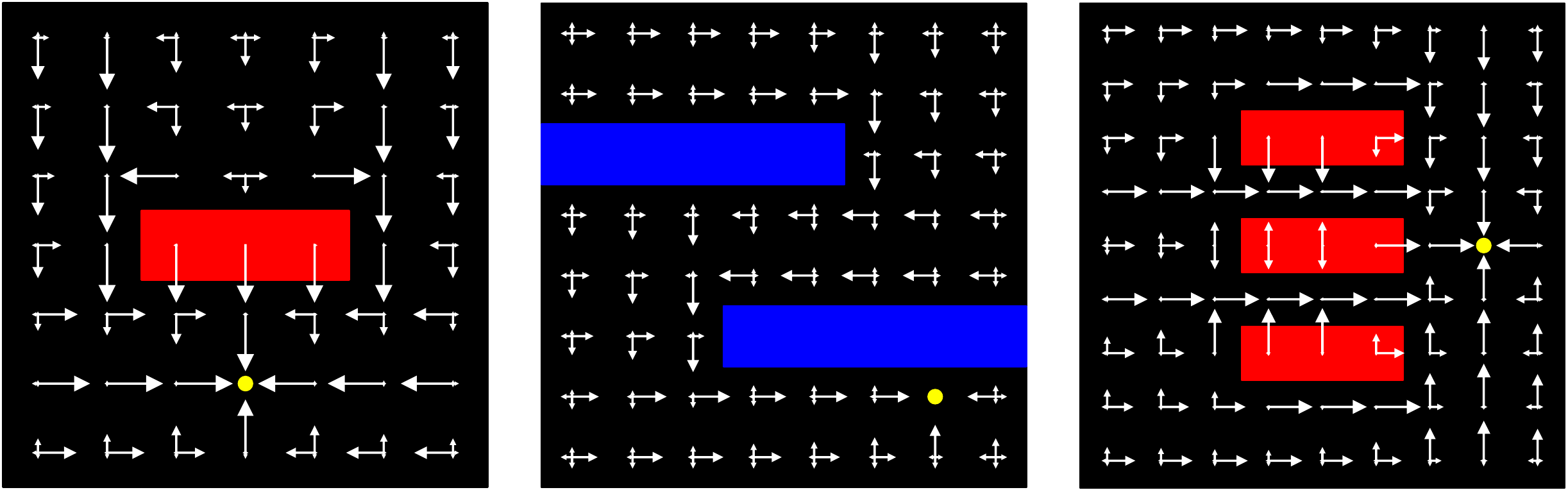}
    \caption{Optimal policies for three different mazes, obtained from the dominant eigenvalue's corresponding left eigenvector of the tilted transition matrix. In these examples, the size of an arrow is proportional to the probability of taking a step in that direction. Blue squares represent hard walls, i.e. the agent is not allowed to step on them. Each step taken by the agent incurs a penalization ($r=-1$). When on a red square, there's a higher penalization ($r=-1.5$) and the agent is allowed to continue its trajectory. The goal state is depicted by the yellow circle, for which there is no penalization ($r=0$) and the agent will be replaced at the initial state, regardless of the action taken.}
    \label{fig:mazes_and_policies}
\end{figure}
Fig.~\ref{fig:mazes_and_policies} shows three examples of mazes and corresponding optimal policies. In the figure we see how the policy can successfully steer the agent toward the goal state, while avoiding \emph{dangerous} states.

\begin{figure}
    \centering
    \includegraphics[width=0.7\linewidth]{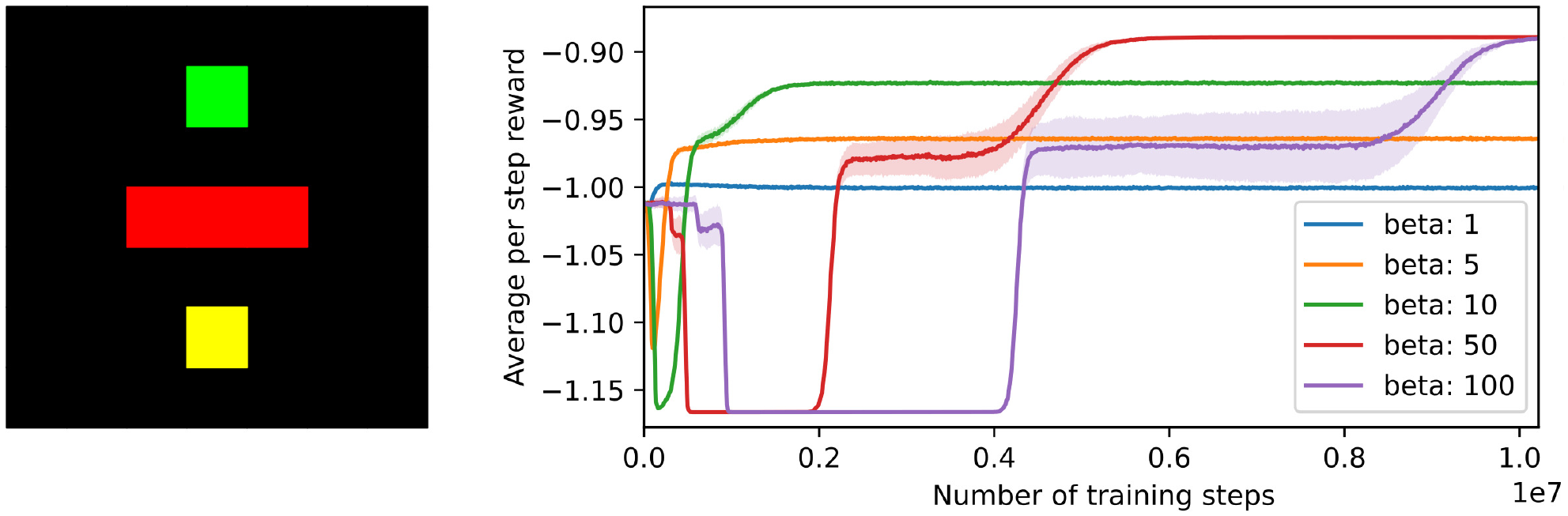}
    \caption{Training evolution of $u$-$\theta$ learning agents for 5 different temperatures as a function of the training progress. Lower temperature agents take longer to converge. Note that the optimal greedy policy is recovered at the lowest temperatures. }
    \label{fig:UThetaLearning_performance_vs_beta}
\end{figure}

\begin{figure}
    \centering
    \includegraphics[width=0.7\linewidth]{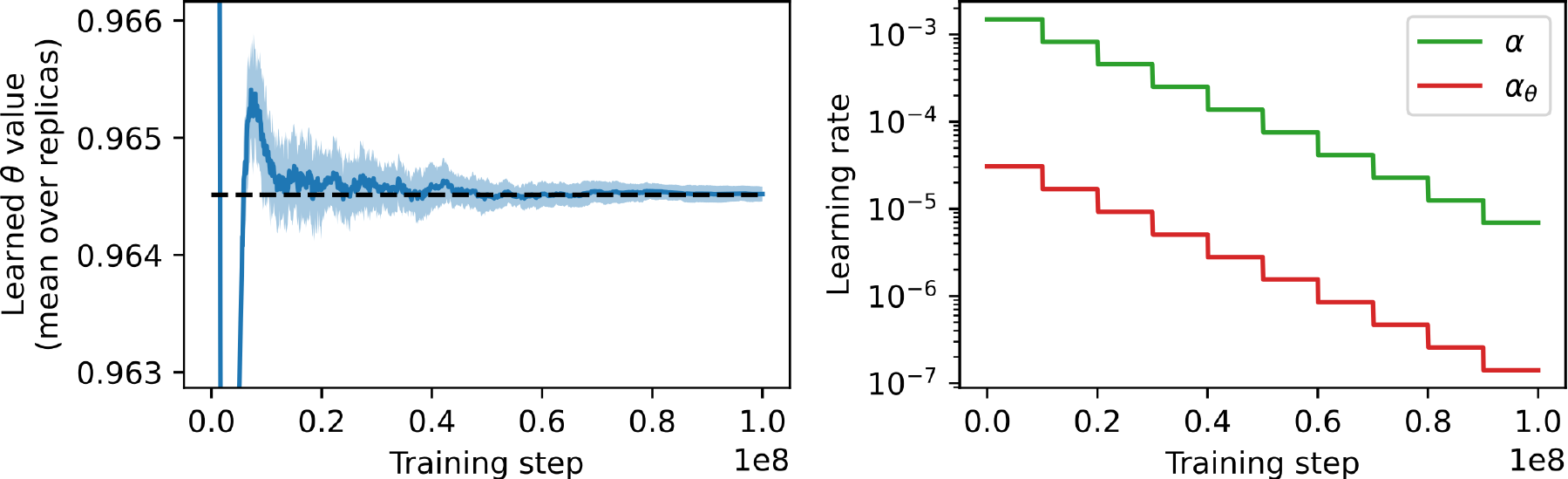}
    \caption{Validation of solution by $u$-$\theta$ learning algorithm. The maze used is the same as in Fig.~\ref{fig:UThetaLearning_performance_vs_beta}, with $\beta = 10$ and trajectory length $N = 1000$ steps.
    Left: Convergence of the $\theta$ parameter learned by the agent towards the target value as computed by the model-based version. The curve plots the mean values over 32 replicas, and the shaded area is the standard deviation.
    Right: Learning rate schedules used to learn the theta parameter.
    }
    \label{fig:utheta_convergence}
\end{figure}

Here we provide some details about the validation of the  model-free version of our method ($u$-$\theta$ learning). The approach consists of a temporal difference method (see Eqs.~\ref{eq:td_update_1} and~\ref{eq:td_update_2} in the main text). Validation of the algorithm has been performed by comparing to the exact solution as computed by dynamic programming.
In Fig.~\ref{fig:UThetaLearning_performance_vs_beta} we show solutions to the displayed maze for several temperatures, as a function of training progress. Fig.~\ref{fig:utheta_convergence} examines the learned parameters and their comparison with dynamic programming.

\end{document}